\documentclass[10pt,twocolumn,letterpaper]{article}

\usepackage[pagenumbers]{cvpr} 

\definecolor{cvprblue}{rgb}{0.21,0.49,0.74}
\usepackage[pagebackref,breaklinks,colorlinks,allcolors=cvprblue]{hyperref}
\usepackage{amsmath,amsfonts,bm}
\usepackage[utf8]{inputenc} 
\usepackage[T1]{fontenc}    
\usepackage{hyperref}       
\usepackage{url}            
\usepackage{booktabs}       
\usepackage{amsfonts}       
\usepackage{nicefrac}       
\usepackage{microtype}      
\usepackage{xcolor}         
\usepackage{graphicx}
\usepackage{caption}
\usepackage{subcaption}
\usepackage{colortbl}
\usepackage{pgfplots}
\usepackage{pgfplotstable}
\usepackage{amsmath}
\usepackage{tabularx}
\usepackage{marvosym}
\usepackage{multirow}

\def\paperID{1401} 
\def\confName{CVPR}
\def\confYear{2025}

\title{MetaFood3D: 3D Food Dataset with Nutrition Values}

\newcommand{\customfootnotetext}[2]{%
  \begingroup
  \renewcommand{\thefootnote}{\fnsymbol{footnote}}%
  \footnotetext[#1]{#2}%
  \endgroup
}

\renewcommand{\thefootnote}{\fnsymbol{footnote}}

\author{
    Yuhao Chen\textsuperscript{2} \and
    Jiangpeng He\textsuperscript{1}\footnotemark[2] \and
    Gautham Vinod\textsuperscript{1}\footnotemark[1] \and
    Siddeshwar Raghavan\textsuperscript{1}\footnotemark[1] \and
    Chris Czarnecki\textsuperscript{2}\footnotemark[1] \and
    Jinge Ma\textsuperscript{1} \and
    Talha Ibn Mahmud\textsuperscript{1} \and
    Bruce Coburn\textsuperscript{1} \and
    Dayou Mao\textsuperscript{2} \and
    Saeejith Nair\textsuperscript{2} \and
    Pengcheng Xi\textsuperscript{2,3} \and
    Alexander Wong\textsuperscript{2} \and
    Edward Delp\textsuperscript{1} \and
    Fengqing Zhu\textsuperscript{1}\\[0.2em]
    \textsuperscript{1}Purdue University \quad
    \textsuperscript{2}University of Waterloo \quad
    \textsuperscript{3}National Research Council Canada
}

\begin{document}
\maketitle

\customfootnotetext{2}{Equal First \& Corresponding Author (\Letter)}
\customfootnotetext{1}{These authors contributed equally to this work for collecting 3D food data}

\begin{abstract}
Food computing is both important and challenging in computer vision (CV). It significantly contributes to the development of CV algorithms due to its frequent presence in datasets across various applications, ranging from classification and instance segmentation to 3D reconstruction. The polymorphic shapes and textures of food, coupled with high variation in forms and vast multimodal information, including language descriptions and nutritional data, make food computing a complex and demanding task for modern CV algorithms. 3D food modeling is a new frontier for addressing food related problems, due to its inherent capability to deal with random camera views and its straightforward representation for calculating food portion size.  However, the primary hurdle in the development of algorithms for food object analysis is the lack of nutrition values in existing 3D datasets. Moreover, in the broader field of 3D research, there is a critical need for domain-specific test datasets. To bridge the gap between general 3D vision and food computing research, we introduce MetaFood3D. This dataset consists of 743 meticulously scanned and labeled 3D food objects across 131 categories, featuring detailed nutrition information, weight, and food codes linked to a comprehensive nutrition database. Our MetaFood3D dataset emphasizes intra-class diversity and includes rich modalities such as textured mesh files, RGB-D videos, and segmentation masks.
Experimental results demonstrate our dataset's strong capabilities in enhancing food portion estimation algorithms, highlight the gap between video captures and 3D scanned data, and showcase the strengths of MetaFood3D in generating synthetic eating occasion data and 3D food objects. The dataset is available at \url{https://lorenz.ecn.purdue.edu/~food3d/}.
\end{abstract}    
\begin{figure*}[h]
\begin{center}
    \includegraphics[width=2. \columnwidth, bb=0 0 703.44 246.96]{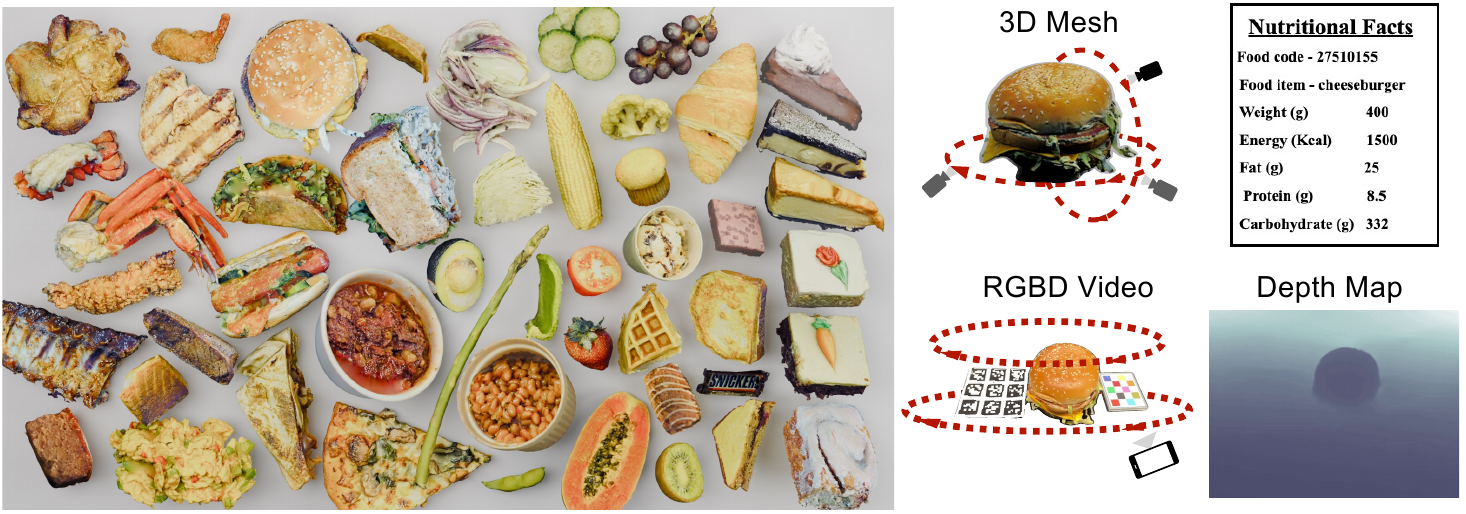}
    \caption{\textbf{MetaFood3D is a real-scan 3D food dataset featuring diverse ready-to-eat 3D textured meshes, 720-degree RGBD video captures, and rich nutrition value annotations.} }
    \label{fig:metafood3d_overview}
    \vspace{-0.5cm}
\end{center}
\end{figure*}

\section{Introduction}
\label{sec:intro}

Food is fundamental to our existence, serving not just as a basic necessity for survival but also as a crucial aspect of our social interactions, where sharing images, videos, and even virtual food experiences in video games is commonplace. Food-related image analysis is crucial for monitoring and improving dietary habits across different age groups, as it enables personalized nutrition interventions, supports early detection of dietary deficiencies, and promotes healthier lifestyles tailored to the specific needs of children, adults, and the elderly. In the field of computer vision, food has played a significant role in advancing algorithms, given its frequent occurrence in both specialized and general datasets for tasks such as classification \cite{gao2022dynamic, he2021online, jiang2019multi, Raghavan_2024}, instance segmentation \cite{lan_foodsam_2023}, and 3D object reconstruction \cite{qian2023magic123}. 

Food data is uniquely complex due to unbalanced classes, intricate textures, hierarchical categorization, and ambiguous shapes.   
Often, food images are taken from close distances, with varying camera angles leading to diverse visual representations. Typical single-view-image depictions fall short of providing comprehensive views, obscuring critical details about ingredients and portions. \textit{E.g.}, an overhead image of a sandwich might display only the bun, while a side view could expose the bun, meat, and toppings in greater detail, highlighting the limitations of single-view image analysis.

Accurate measurement is crucial for various food-related tasks, especially under the context of precise dietary assessment, which can serve as a valuable digital biomarker, offering a quantitative and objective measure of an individual's nutritional intake and its potential impact on their health status. A significant challenge in dietary assessment is to accurately estimate portion sizes from food images \cite{volume-survel-2021}. Various approaches have been developed to tackle this problem, including image based regression \cite{nutrition5k}, regression on segmentation masks \cite{segmentation_based, segmentation-based}, mapping to handcrafted 3D shape templates \cite{jia2023estimating_bowl}, 3D reconstruction from multiple images \cite{3d_reconstruction}, and utilizing depth information \cite{graikos2020depth}. However, the lack of 3D information for individual food object leads to inaccuracies and challenges in generalization. Even with depth data, accurately representing empty spaces beneath food objects remains a challenge, as foods on a plate can exhibit a wide range of 6D poses and stacking relationships. 

Recent advancements in 3D vision algorithms, particularly in novel view synthesis \cite{mildenhall2021nerf}, surface reconstruction \cite{wang2021neus}, and 3D object generation \cite{lin2023magic3d}, indicate a promising direction for overcoming these issues. Utilizing 3D methodologies in food-related research offers inherent advantages, such as mitigating challenges posed by varied camera views through novel view synthesis or rendering from learned geometries. These approaches can facilitate the direct computation of food volume per food item for dietary studies, making the process more precise, straightforward, and explainable compared to existing methods. However, at this stage, the main obstacle to applying these 3D algorithms to food-related tasks is the lack of well constructed food datasets.

Many generic large-scale 3D datasets \cite{deitke2023objaverse, wu2023omniobject3d, google_scanned_objects} have recently been released, fueling the development of 3D vision algorithms \cite{shi2023mvdream, liu2023syncdreamer}. Yet, there is a notable scarcity of food-specific datasets to train and evaluate 3D algorithms on food-related tasks. Existing 3D datasets with food generally lack dietary annotations such as weight, calories, and other nutrition values, which is crucial for developing 3D or image-based dietary assessment algorithms. Furthermore, there is a shortage of benchmark 3D food datasets featuring diverse intra-class variation. For instance, the OmniObject3D dataset \cite{wu2023omniobject3d} includes 2,837 food objects, but the selection of its food instances fails to emphasize the appearance variations within each food category. Many food items in OmniObject3D, such as lemons, exhibit similar appearances and geometries within the same category. 

To bridge the gap between general 3D vision and food computing, and to provide a unique benchmark for both general and food-specific downstream tasks, our dataset MetaFood3D (as shown in Figure \ref{fig:metafood3d_overview}) endeavors to develop a food-specific 3D dataset that advances dietary analysis from 2D to 3D. MetaFood3D includes a total of 743 3D food objects in 131 food categories. Each food object in the dataset is meticulously labeled with detailed nutrition information, weight, and food codes linked to a comprehensive nutrition database~\cite{Montville2013FNDDS}. We emphasize intra-class diversity by collecting foods with varying appearances and nutritional information. Beyond nutritional facts, our dataset includes rich modalities such as textured mesh files, RGB-D videos, and segmentation masks. Additionally, the dataset incorporates hierarchical relationships characterized by specifying sub-food-categories, known as food items, within general food categories, facilitating tasks related to fine-grained classification. Finally, we establish baselines for nutrition estimation, perception, reconstruction, and generation tasks. Our experiments demonstrate that our dataset has significant potential for improving performance and highlight the challenging gap between video captures and 3D scanned data. Furthermore, we show the potential of our dataset for high-quality data generation, simulation, and augmentation by presenting high-quality visual results.

\section{Related Work}
\label{sec:realted_work}
\begin{table*}[]
\scalebox{.87}{
\begin{tabular}{c|ccccccccc}
    \toprule
     & Multiview/video & Depth & Inst Mask & Mesh & Size Calibration & Nutrition & Food categories & Samples\\
    \midrule
    \underline{Food Specific Datasets} &  &  &  &  &  &  & &  \\
    Food101 \cite{food101} (2D) &  &  &  &  &  &  & 101 & 101,000   \\    
    Food2K \cite{food_rec_2023} (2D) &  &  & \checkmark &  &  &  & 2,000 & 1 Million   \\
    ECUSTFD \cite{ecustfd_food_dataset} (2D) &  &  &  &  & \checkmark & \checkmark & 19 & 2,978   \\
    Nutrition5K \cite{nutrition5k} (2D) & \checkmark & \checkmark &  &  & \checkmark & \checkmark & 250 & 5,006  \\
    NutritionVerse3D \cite{taiNutritionVerse3D3DFood2023} & \checkmark &  & \checkmark & \checkmark &  & \checkmark & 54 & 105  \\

    \midrule
    \underline{  Generic 3D Datasets  } &  &  &  &  &  &  & &  &  \\
    GSO \cite{google_scanned_objects} &  &  & \checkmark & \checkmark & \checkmark &  & 0 & 0  \\
    CO3D \cite{co3d} & \checkmark & \checkmark & \checkmark &  & \checkmark &  & 10 & 5,077  \\
    OmniObject3D \cite{wu2023omniobject3d} & \checkmark &  & \checkmark & \checkmark & \checkmark &  & 85 & 2,837  \\

    \midrule
    
    Ours & \textbf{\checkmark} & \textbf{\checkmark} & \textbf{\checkmark} & \textbf{\checkmark} & \textbf{\checkmark} & \textbf{\checkmark} & \textbf{131} & 743  \\
    \bottomrule
\end{tabular}
}
\caption{\textbf{Public Datasets with Real-world Food Objects.} ``Samples" represents the total number of food data samples in the dataset. Note that we exclude food toys in GSO. }
\label{tab:food_specifics}
\end{table*}


In this section, we provide detailed reviews of related food and 3D object datasets and a brief review of relevant downstream tasks. The features of these datasets are summarized in Table \ref{tab:food_specifics}.

\textbf{Food Datasets} are primarily developed to answer key questions in food computing: ``What is the food in the image?'', ``What is the portion size?'', and ``What is the nutritional content of the food?''. While numerous food classification datasets exist, ranging from the classic Food-101 dataset \cite{food101} to the latest Food2K dataset \cite{food_rec_2023}, datasets for portion estimation or macro-nutrient estimation are significantly fewer. 
This scarcity is due to the complexity and labor-intensiveness of collecting multi-modal data with physical food object references. Numerous efforts have been undertaken to mitigate the need for gathering data on physical objects. These include leveraging images and metadata from recipe websites \cite{pic2kcal} or creating synthetic data by pasting image textures onto predefined geometries \cite{food-volume-single-view-syn}. However, these approaches have fundamental flaws, as the relationship between the food appearance and the food weight is not validated by real food items.
Despite various proposals for ground-referenced food portion estimation datasets in existing literature \cite{review_2020, tahir_comprehensive_2021, wang_review_2022, review_2024}, only three datasets that include nutrition values are publicly available: ECUSTFD \cite{ecustfd_food_dataset}, Nutrition5K \cite{nutrition5k}, and NutritionVerse3D \cite{taiNutritionVerse3D3DFood2023}. The ECUSTFD dataset contains no geometry information. In the Nutrition5K dataset, food items are mixed together without segmentation masks, making it infeasible to perform nutrition and geometric modeling for individual food items. The NutritionVerse3D dataset, which includes models from FoodVerse \cite{tai2022foodverse}, is small-scale, containing 105 3D food models across 42 unique food types. The food items are not calibrated in size and the selection of food types appears to be random and imbalanced.

\textbf{3D Object Datasets} focus either on synthetic objects created by humans or on real-world objects that are manually scanned. Synthetic object datasets, such as ShapeNet \cite{chang2015shapenet} and Objaverse \cite{deitke2023objaverse}, 
are unsuitable for dietary assessment applications due to their artistic object appearances and non-referenced scales.
Real-world scanned objects offer realistic appearances and geometry, but many real-world 3D object datasets primarily focus on non-perishable commercial household items, including Google Scanned Objects (GSO) \cite{google_scanned_objects}, CO3D \cite{co3d}, YCB Objects \cite{calli2015ycb}, AKB-48 \cite{akb48}, and MetaGraspNetV2 \cite{metagraspnetv2}. Some real-world scanned object datasets do include food items, but they often suffer from limitations such as a small number of food categories \cite{co3d}. Additionally, the selection of food items is often random and does not reflect the distribution of commonly eaten foods, leading to bias in dietary assessment \cite{wu2023omniobject3d}.




\textbf{Food Data Analysis for Dietary Assessment.}
Existing food portion and nutrition value estimation methods can be classified into four main categories: stereo-based~\cite{Puri2009Stereo, Dehais2017TwoView}, depth-based~\cite{Lo2019Depth, Fang2016Depth}, model-based~\cite{Xu2013ModelBased, jia2014modelbased}, and neural network-based methods~\cite{he2020multitask, Shao2021EnergyDensity, Ma2023DensityMapSumming, Vinod2022EnergyDensityDepth, he2021end, nutrition5k, Shao2023VoxelReconstruction}. 
Recently, 3D model-based methods \cite{vinod2024Model3D, jinge2024MPF3D} have demonstrated the importance of 3D models in food portion estimation by outperforming many existing methods. 

\textbf{3D Point Cloud Perception.} This task seeks to classify point cloud data composed of a set of 3D coordinates. PointNet \cite{qi2017pointnet} was first proposed to directly process unordered raw point cloud sets.  PointNet then led to the development of new models \cite{qi2017pointnet++, wang2019dynamic,xu2021paconv,ma2022rethinking}. Due to the characteristics of real-world point cloud data, robustness is crucial in 3D point cloud perception. Previous works \cite{ahmadyan2021objectron, reizenstein2021common, ren2022benchmarking,taghanaki2020robustpointset} have studied the robustness of models on point cloud data from different domains and standardized corrupted dataset. 

\textbf{Novel View Synthesis and 3D Mesh Reconstruction.} Novel view synthesis aims to generate high-quality images from new perspectives given only a few training images. Neural Radiance Fields (NeRF) \cite{mildenhall2021nerf} addresses this problem by training a multilayer perceptron (MLP) network to predict the color values and densities of locations in space. Recent advancements have tackled issues related to aliasing, quality, and efficiency \cite{barron2021mipnerf, mueller2022instant, gaussiansplatting, nerfstudio}.
3D mesh reconstruction aims to recreate the mesh of an object. Traditional methods like Structure from Motion (SfM) \cite{schoenberger2016sfm} achieve this by determining the camera pose associated with each image. Recent approaches leverage the success of volume rendering in novel view synthesis \cite{wang2021neus, li2023neuralangelo, 2d_gaussian} or employ Neural Signed Distance Fields \cite{nvdiffrec}.


\textbf{3D Generation.} With advancements in novel view synthesis and generative models \cite{ldm}, numerous text-to-3D generation methods have emerged in the past year \cite{text-2-3D-survey}. A typical pipeline involves leveraging diffusion models to generate multi-view images of an object, which are then utilized in 3D reconstruction methods to create the 3D model \cite{shi2023mvdream, long2023wonder3d}. Other approaches focus on learning Neural Signed Distance Fields to achieve 3D generation \cite{gao2022get3d} . 



\begin{figure*}[t]
  \centering
    \includegraphics[width=2.0\columnwidth]{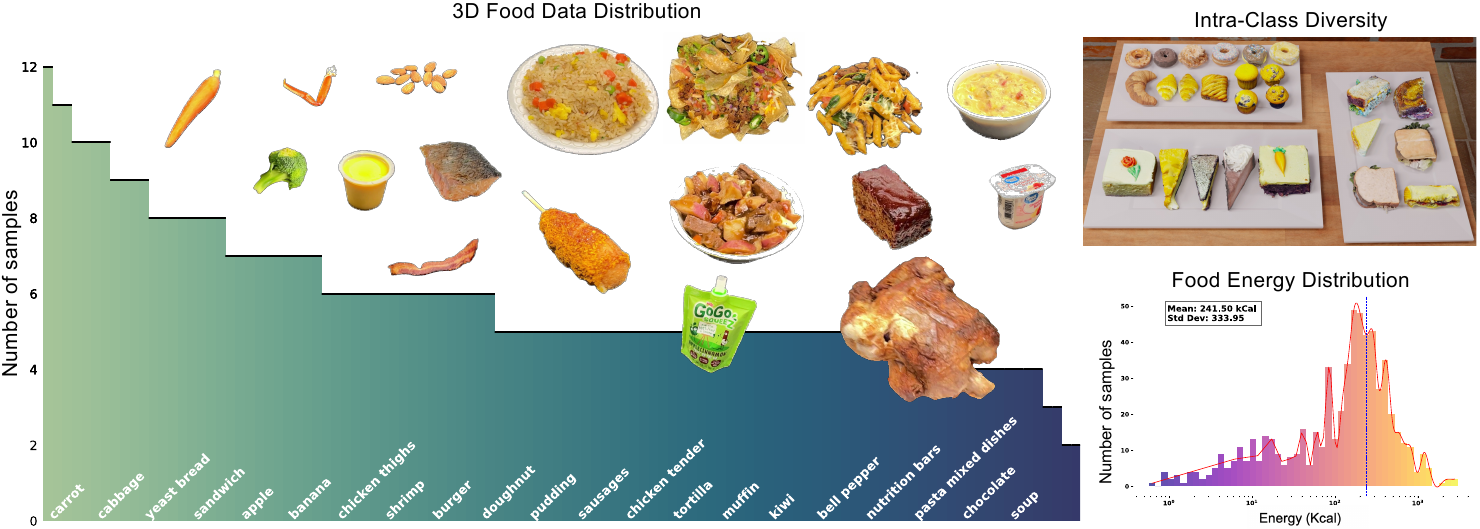}
    \caption{\textbf{The distribution of MetaFood3D}, which includes 131 mostly consumed food categories with high intra-class diversity, a total of 220 unique food items, each matched to a unique food code, and 743 single food objects in total with each containing nutrition values annotations.}
    \label{fig:object_samples}
\end{figure*}

 \section{Dataset}
\label{sec:dataset}
The selection of food objects and their multimodal labels in the MetaFood3D dataset is designed to support dietary assessment applications, which involves identifying various foods in images and estimating portion sizes and nutritional values using RGB and/or depth sensors from diverse camera angles. To accurately reflect these use cases, we first carefully selected food items and their variations based on real-world food consumption patterns, as detailed in the \textbf{Food Objects Selection} paragraph. Second, we curated the modalities and labels to capture the relevant characteristics of real-world dietary assessment data, as described in the \textbf{Data Collection} and \textbf{Annotation} paragraph. Figure \ref{fig:object_samples} provides an overview of MetaFood3D, illustrating the distribution of data and energy content across food objects, as well as the intra-class variance of the collected food objects. 

\textbf{Food Objects Selection.} Identifying which food objects to collect is challenging due to the vast number of food categories and the significant appearance variations even within the same category. For example, apples could be broadly categorized as fruit, but they also come in different varieties, colors, shapes, and sizes, and can be used in diverse preparations like apple pies. Determining the appropriate level of class granularity poses another challenge—should we classify broadly as "fruit," more specifically as "apple," or even further as "Fuji apple"? To address these challenges, we consulted nutrition experts and referenced an established food list from the VIPER-FoodNet (VFN) dataset~\cite{mao2021visual}. The VFN dataset, derived from the What We Eat in America (WWEIA) database\footnote{https://data.nal.usda.gov/dataset/what-we-eat-america-wweia-database}, provides a comprehensive overview of the American diet. It has been widely used in food computing tasks, such as long-tailed learning~\cite{he_long-tailed_2023}, continual learning~\cite{Raghavan_2024}, personalized classification~\cite{pan2023personalized}, and multimodal learning~\cite{pan2024fmifood}. To enhance categorical diversity, we expanded the original 74 food categories from the VFN dataset by incorporating 57 additional categories based on data from the National Health and Nutrition Examination Survey (NHANES)~\cite{lin2022differences}, resulting in a total of 131 food categories in the MetaFood3D dataset. This expansion not only increases the dataset's coverage but also enhances its cultural diversity, as NHANES includes foods from various cultural backgrounds (e.g., sushi from Asian cuisines), making our dataset more representative of the multicultural nature of contemporary American dietary patterns.
One key enhancement of our dataset over the VFN dataset is the increased granularity of food code matching. While VFN matches each food category with a single general 8-digit food code from the Food and Nutrient Database for Dietary Studies (FNDDS)~\cite{Montville2013FNDDS}, resulting in only 74 food codes for 74 food categories, our approach provides a more granular mapping. Specifically, we assign FNDDS food codes at two levels including both food categories and individual food items. This hierarchical structure includes 131 distinct food codes for food categories and 743 unique food codes for specific food items. For example, within the "Pie" category, we include specific items like "Pie, chocolate cream," "Pie, pecan," "Pie, apple," and "Pie, lemon," each with their respective FNDDS codes. This detailed matching allows for a more accurate representation of diverse food items, acknowledging their unique ingredients and nutritional profiles. By providing this level of detail, our 3D food dataset enables more precise dietary analysis and the development of sophisticated computer vision algorithms capable of distinguishing between different food items within a category. Our fine-grained categorization results in a total of 220 food items, each with a unique FNDDS code, forming the foundation of our 3D data collection process. Including various food items within each category allows our collected 3D models to capture intra-category visual and geometric diversity, enhancing the accuracy of algorithms for dietary assessments. 
When balancing category diversity against within-category diversity, we chose to prioritize expanding the range of food categories. This decision stems from our belief that generative models have significant potential for data augmentation, enabling scalable expansion of the dataset beyond what manual collection alone can achieve. By focusing on category diversity, our 3D food models can serve as prototypes that can be further enhanced by leveraging internet-scale priors--which would be more challenging if we concentrated solely on within-category variations.


\textbf{Data Collection.} We prioritize sourcing real-world food objects from restaurants and ready-to-eat or frozen foods from grocery stores. For food that are difficult to source, we prepare them from raw ingredients such as peanut butter and jelly sandwich.
Besides leveraging both the food category and food item categorization, we also enhance intra-class diversity during the data collection step by employing various food sourcing strategies. These include sourcing food from different restaurants, stores, or locations; selecting diverse flavors, brands, breeds, or forms; cutting, peeling, or unwrapping the food; and preparing the food with different ingredients. These strategies ensure that our dataset captures a wide range of appearances and geometries for each food category.
Our 3D data collection follows a similar approach to OmniObject3D \cite{wu2023omniobject3d} and NutritionVerse3D \cite{taiNutritionVerse3D3DFood2023}. The food object is placed on a turntable and scanned by a 3D scanner, the Revopoint POP 2\footnote{https://www.revopoint3d.com/pages/face-3d-scanner-pop2}, which is positioned statically on a tripod. We then record the food's weight and nutrition value. For most objects, keypoint tracking provided by RevoScan software \cite{revoscan} is sufficient to obtain a 360-degree point-cloud capture of the food object. If the scan is not successful, we manually turn the object.
Unlike OmniObject3D \cite{wu2023omniobject3d}, which captures a 360\textsuperscript{$\circ$} range, we perform a 720\textsuperscript{$\circ$} RGBD video capture by rotating the object twice in a spiral motion, ending with an overhead capture. This approach ensures that we capture the most likely camera angles from typical smartphone users. If the food object can be flipped (e.g., a bowl of beef stew cannot be flipped), we flip the object and repeat the RGBD video data capture process to capture the underside of non-fluid objects. The depth measurement is obtained using an iPhone App called Record3D \cite{Record3D3DVideos}. To ensure precise scale and color measurements, we use calibration fiducial markers~\cite{xu2012image-FM} for both camera angle and color calibration. Details of our data collection pipeline can be found in our supplementary materials.

\textbf{Annotation.} After collecting the 3D food objects, we perform a series of postprocessing steps and annotate each food object. One of our unique contributions is the annotation of weight and nutrition facts for each food object, which is crucial for food data and dietary assessment tasks. During the data collection process, we record the weight $w_i$ (in grams) of each food object $i$. By leveraging the food code associated with each object, we obtain the nutrient value density $d_i$, which represents the nutrient content per 100 grams of the food item. The nutrient value density is typically expressed as a vector $d_i = [e_i, p_i, c_i, f_i]$, where $e_i$, $p_i$, $c_i$, and $f_i$ denote the energy (in kilocalories), protein (in grams), carbohydrates (in grams), and fat (in grams) per 100 grams of food item $i$, respectively.
Given the weight $w_i$ and nutrient value density $d_i$, following~\cite{he_long-tailed_2023, lin2023integration}, we can determine the total nutrient content $n_i$ for the specific quantity of food object $i$ in our dataset with $n_i = \frac{w_i}{100} \cdot d_i$.  The inclusion of weight and nutrition values enables researchers to develop and evaluate algorithms for precise dietary assessment and nutrient estimation. Similarly, as in~\cite{wu2023omniobject3d}, we also generate data to support various general 3D vision research topics such as point cloud analysis, neural radiance fields, and 3D generation. This includes rendering object-centric and photo-realistic multi-view images using Blender \cite{blender} with accurate camera poses, generating depth and normal maps, and sampling multi-resolution point clouds from each 3D model. Additionally, for the collected RGBD videos, we provide uniformly sampled video frames with corresponding segmentation masks and depth information. The segmentation masks are generated based on GroundingDINO~\cite{grounding_dino}, Segment Anything Models (SAM)~\cite{kirillov_segment_2023} and Cutie~\cite{cutie}.

Overall, We collected 743 food objects with 131 food categories. Each food object in our dataset includes the following labels: a scanned 3D object mesh with texture, RGBD video capture of the food both in a standard pose and flipped (if applicable), depth images and masks corresponding to the RGBD video captures, FNDDS food code, nutrition value (energy, protein, carbohydrates, fat), weight value, Blender-rendered frames with normal and depth images, camera parameters used for rendering, and fiducial marker (with known physical dimensions) used in the video capture.


\section{Experimental Results}
\label{sec:experiment}
In this section, we demonstrate the usage of the MetaFood3D dataset in four downstream tasks: 3D food perception (Section~\ref{sub_sec:perception}), novel view synthesis and 3D reconstruction (Section~\ref{subsec: novel view and reconstruction}), 3D food generation and rendering (Section~\ref{subsec: 3d food generation and rendering}), and food portion size estimation (Section~\ref{subsec: portion estimation}).
The implementation details of all experiments are available in Supplementary Materials. 

\label{subsec: 3d food perception}
\definecolor{mintblue}{RGB}{173,216,230}
\definecolor{skyblue}{RGB}{78,182,255}
\definecolor{lightred}{RGB}{253,226,228}
\definecolor{lightpurple}{RGB}{223,230,253}


\begin{table}[ht]
    \centering
    \resizebox{1.0\columnwidth}{!}{
    \begin{tabular}{l|cc|cc}
        \toprule
        \textbf{} & {OA$_{\text{Uniform}}$ $\uparrow$} & {OA$_{\text{Diverse}}$ $\uparrow$} & {OA$_{\text{Clean}}$ $\uparrow$} & {mCE} $\downarrow$ \\
        \midrule
        DGCNN \cite{wang2019dynamic} & \cellcolor{mintblue!10}0.862 & \cellcolor{mintblue!10}0.196  & \cellcolor{skyblue!20}0.754  & \cellcolor{skyblue!40}1.000  \\
        PointNet \cite{qi2017pointnet} & \cellcolor{mintblue!0}0.822 & \cellcolor{mintblue!0}0.181 & \cellcolor{skyblue!0}0.698  & \cellcolor{skyblue!0}1.210  \\
        PointNet++ \cite{qi2017pointnet++} & \cellcolor{mintblue!30}0.893 & \cellcolor{mintblue!40}0.208  &  \cellcolor{skyblue!90}\textbf{0.788} & \cellcolor{skyblue!90}\textbf{0.912}\\
        SimpleView \cite{goyal2021revisiting} & \cellcolor{mintblue!90}\textbf{0.919} & \cellcolor{mintblue!60}0.223  & \cellcolor{skyblue!70}0.753  & \cellcolor{skyblue!50}0.992  \\
        GDANet \cite{xu2021learning} & \cellcolor{mintblue!50}0.903 & \cellcolor{mintblue!30}0.195 & \cellcolor{skyblue!50}0.766 & \cellcolor{skyblue!80}\underline{0.935}  \\
        PAConv \cite{xu2021paconv} & \cellcolor{mintblue!20}0.892 & \cellcolor{mintblue!20}0.203  & \cellcolor{skyblue!10}0.730  & \cellcolor{skyblue!10}1.036 \\
        CurveNet \cite{xiang2021walk} & \cellcolor{mintblue!60}0.906 & \cellcolor{mintblue!70}0.228 & \cellcolor{skyblue!60}0.763  & \cellcolor{skyblue!60}0.966 \\
        RPC \cite{ren2022benchmarking} & \cellcolor{mintblue!40}0.900 & \cellcolor{mintblue!50}0.206 & \cellcolor{skyblue!40}\underline{0.771}  & \cellcolor{skyblue!70}0.959 \\
        PointMLP \cite{ma2022rethinking} & \cellcolor{mintblue!70}0.912 & \cellcolor{mintblue!90}\underline{0.245} & \cellcolor{skyblue!80}0.770 & \cellcolor{skyblue!20}1.033 \\
        Point-BERT \cite{yu2022point} & \cellcolor{mintblue!80}\underline{0.914} & \cellcolor{mintblue!80}\textbf{0.246}  & \cellcolor{skyblue!30}0.754  & \cellcolor{skyblue!30}1.013\\
        \bottomrule
    \end{tabular}
    }
    \caption{\textbf{Robustness Analysis} on Intra-class Diversity and Point Clouds Corruption}
    \label{tab:3d perception-1}
\end{table}

\subsection{3D Food Perception}
\label{sub_sec:perception}

\textbf{Intra-class Diversity of Food Shapes}: Food objects in real-world settings are often processed into various shapes, such as whole fruits versus sliced fruits or a single nut compared to multiple nuts in a bowl. 
To demonstrate the impact of shape diversity on 3D perception algorithms, we select and train 10 existing methods on OmniObject3D and evaluate their performance on both OmniObject3D (OA$_{\text{Uniform}}$) and MetaFood3D (OA$_{\text{Diverse}}$) using shared food categories. Overall Accuracy (OA) is used to measure the models' robustness against diverse point cloud shapes.
Table~\ref{tab:3d perception-1} shows that OA$_{\text{Diverse}}$ was generally 70\% lower than OA$_{\text{Uniform}}$, indicating that models trained with relatively uniform shapes achieved significantly degraded performance on diverse-shaped food test set. This finding highlights the importance of incorporating shape diversity in 3D food datasets, a key strength of MetaFood3D, ensuring the robustness and generalizability of 3D perception algorithms in real-world applications.

\textbf{Corruption in Point Clouds.} Real-world 3D point clouds of food items can be affected by various types of corruptions, such as noise, missing points, or scaling issues, arising from factors such as sensor limitations, or variations in scanning conditions. To evaluate the robustness of 3D perception models under these corruptions, we created MetaFood3D-C by modifying MetaFood3D with common corruptions described in \cite{ren2022benchmarking}. OA$_{\text{Clean}}$ represents the overall accuracy on the clean MetaFood3D test dataset. The mean Corruption Error (mCE) \cite{ren2022benchmarking} corresponds to the models tested on the MetaFood3D-C to assess their performance in the presence of real-world corruptions. As shown in Table \ref{tab:3d perception-1}, PointNet++ and GDANet demonstrate the best robustness on average against various corruptions. The full results can be found in the Supplementary Materials.

\subsection{Novel View Synthesis and 3D Reconstruction}
\label{subsec: novel view and reconstruction}

\begin{figure}[h]
    \centering
    \begin{subfigure}[t]{0.49\columnwidth}
        \centering
        \includegraphics[width=\columnwidth, bb=0 0 1264 845]{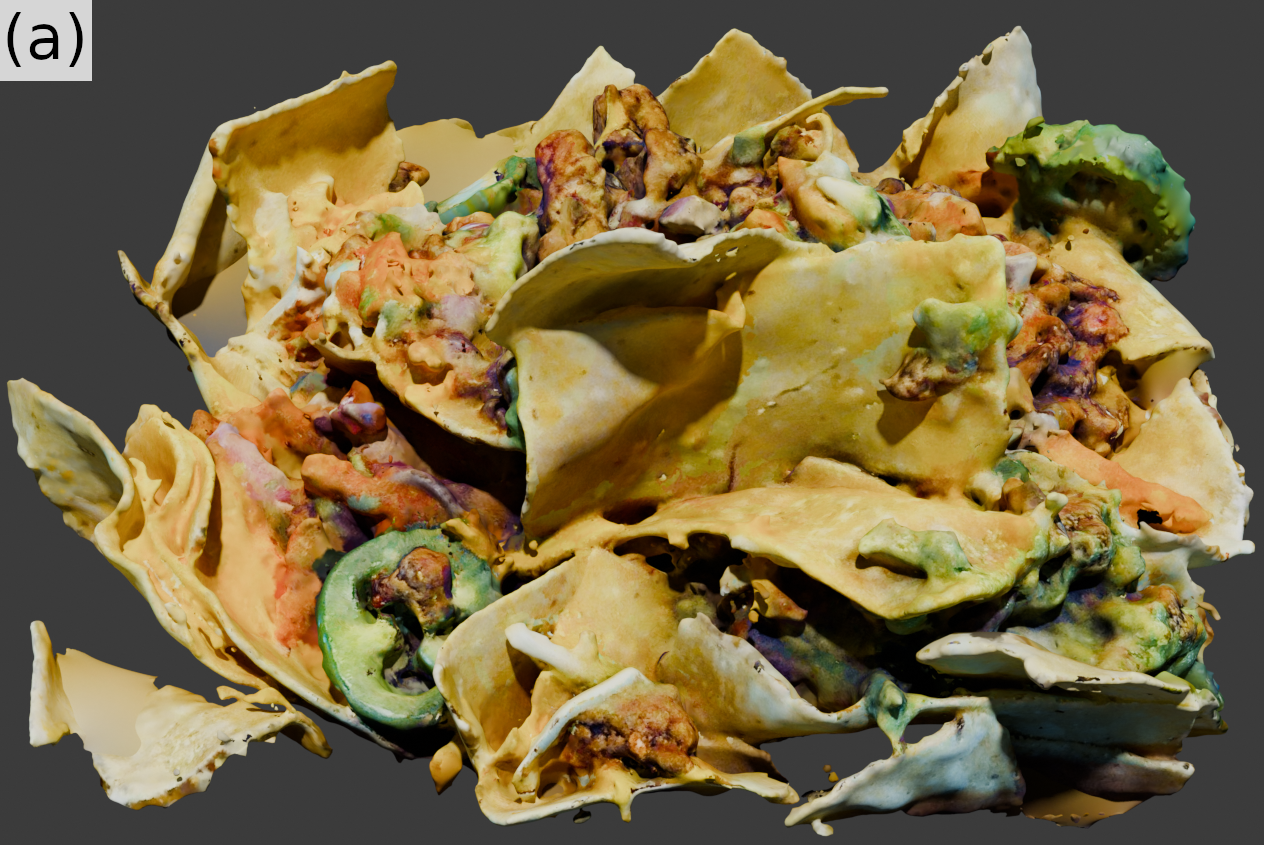}
        \phantomcaption
        \label{fig:nerfacto-nacho-a}
    \end{subfigure}
    \begin{subfigure}[t]{0.49\columnwidth}
        \centering
        \includegraphics[width=\columnwidth,bb=0 0 1264 845]{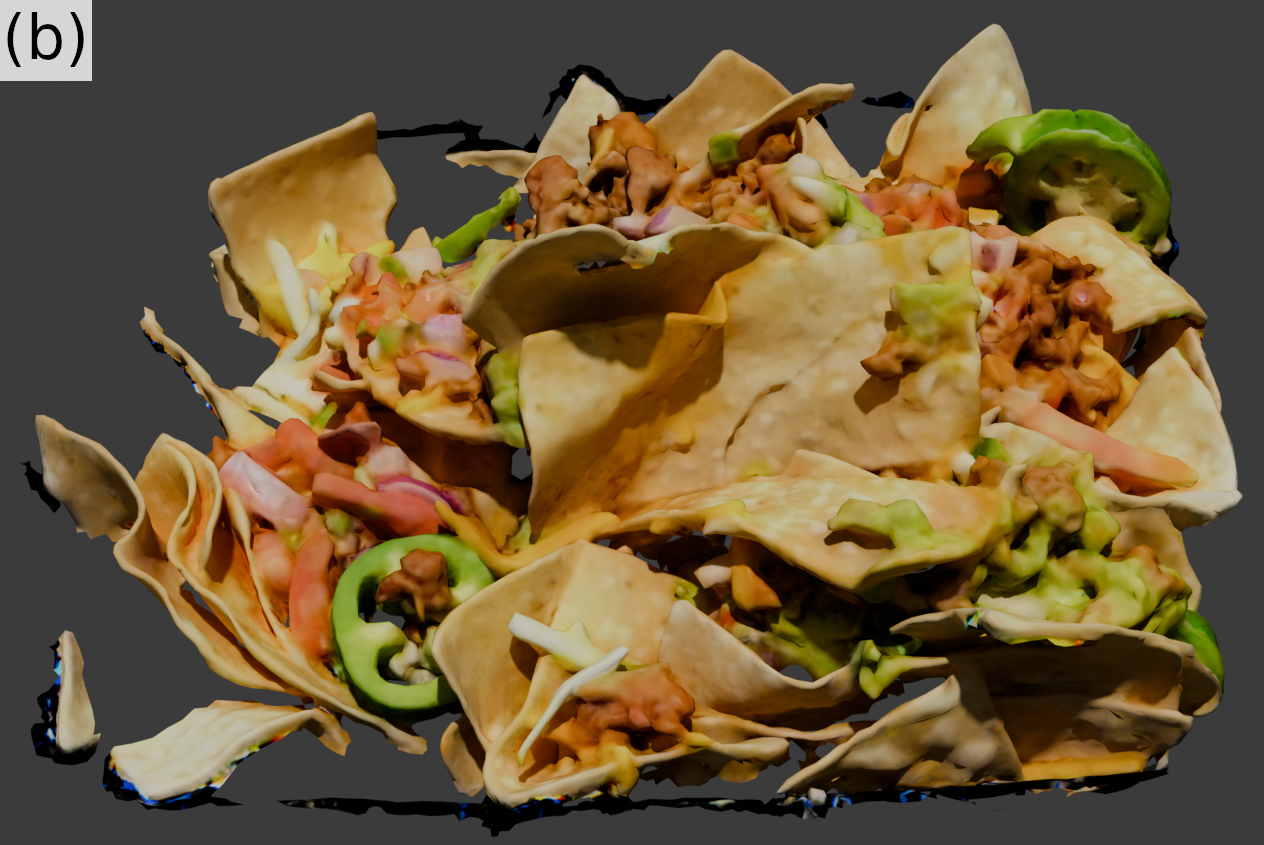}
        \phantomcaption
        \label{fig:nerfacto-nacho-b}
    \end{subfigure}
    
    \vspace{-0.4cm}
    \begin{subfigure}[t]{0.49\columnwidth}
        \centering
        \includegraphics[width=\columnwidth,bb=0 0 1263 843]{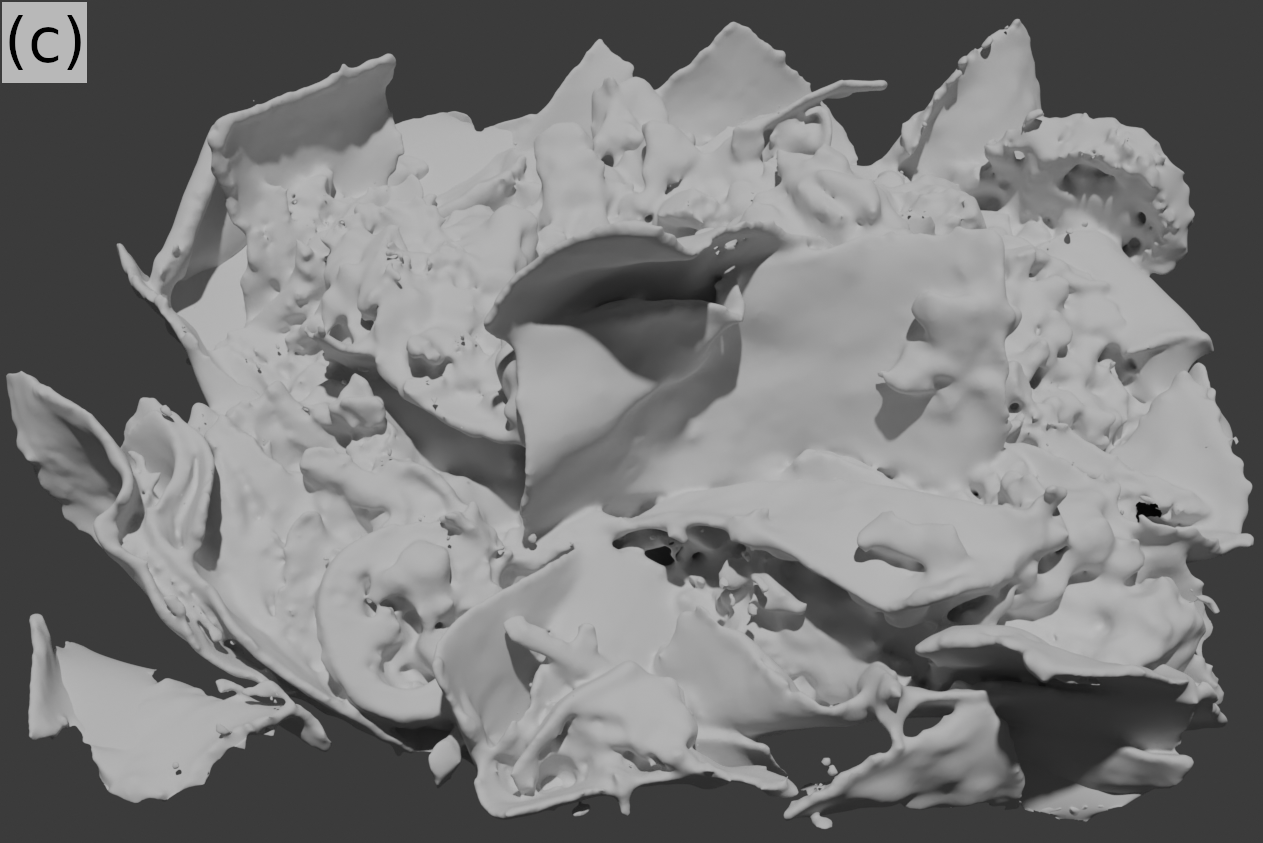}
        \phantomcaption
        \label{fig:nerfacto-nacho-c}
    \end{subfigure}
    \begin{subfigure}[t]{0.49\columnwidth}
        \centering
        \includegraphics[width=\columnwidth,bb=0 0 1263 843]{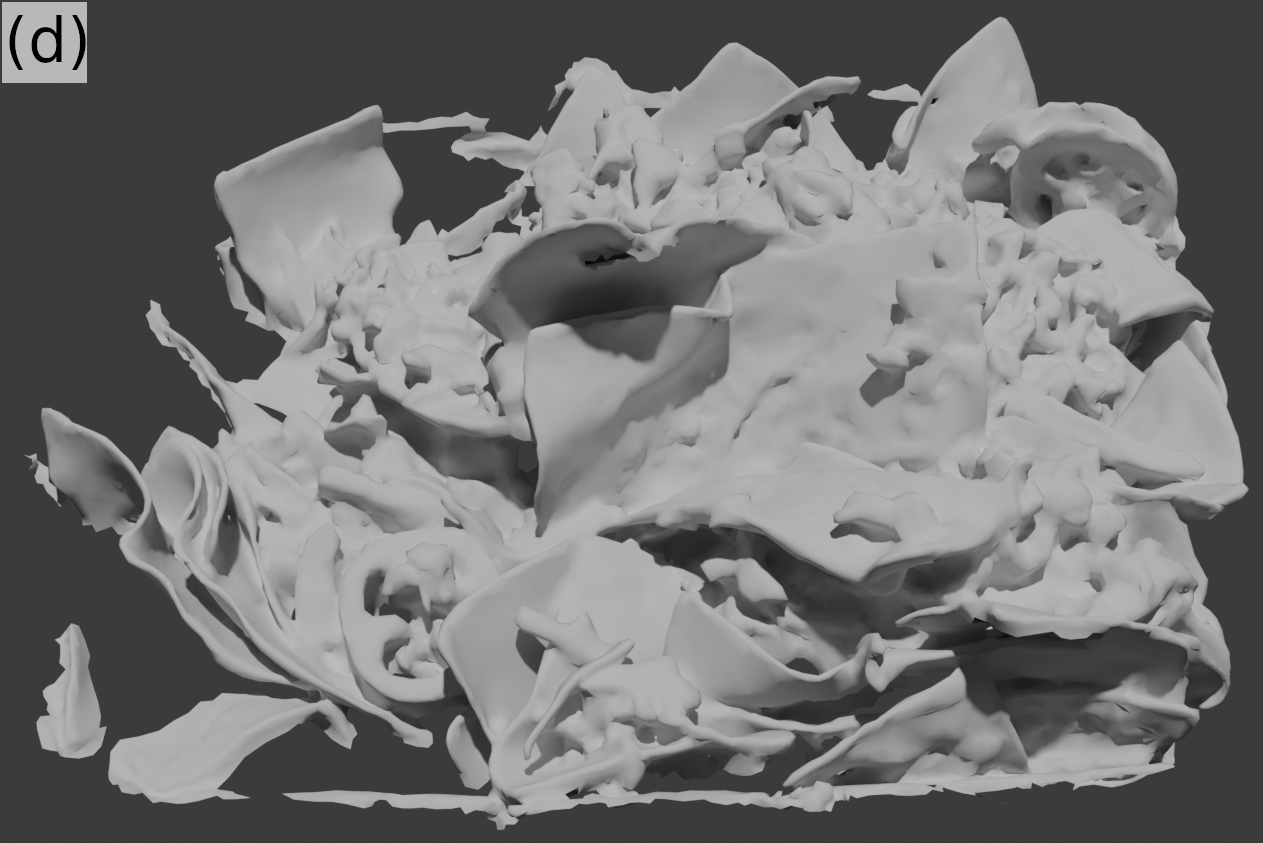}
        \phantomcaption
        \label{fig:nerfacto-nacho-d}
    \end{subfigure}
    \vspace{-0.4cm}
    \caption{\textbf{Reconstructed Mesh:} \textbf{(a)} Ground-truth textured 3D mesh of a complex food item (nachos). \textbf{(b)} A textured 3D mesh of the same food item (nachos) reconstructed from video using Nerfacto. \textbf{(c)} and \textbf{(d)} are mesh-only views of the ground truth and the reconstructed model respectively.}
    \label{fig:nerfacto-nacho}
\end{figure}

\begin{table}[h]
    \centering
    \resizebox{1.0\columnwidth}{!}{
    \begin{tabular}{c||cccc}
        \toprule
        Method & Input & PSNR ($\uparrow$) & SSIM ($\uparrow$) & LPIPS ($\downarrow$) \\
        \midrule
        \multirow{2}{*}{Nerfacto \cite{nerfstudio}} & Render & 20.08 & 0.9219 & 0.0887  \\
           & Video & 22.74 & 0.9633 & 0.0712  \\
           \hline
        \multirow{2}{*}{Nerfacto  (masked)} & Render & 20.06 & 0.9225 & 0.0901  \\
         & Video & 9.10 & 0.0586 & 1.0644   \\
         \hline
        \multirow{2}{*}{3DGS \cite{gaussiansplatting}} & Render & 43.86 & 0.8574 & 0.0868  \\
         & Video & 37.83 & 0.9897 & 0.0114  \\
        \hline
    \end{tabular}
    }
    \caption{\textbf{Novel view synthesis results} on 131 categories. ``Render'' represents rendered Blender data from ground truth meshes and ``Video'' represents captured video data.}
    \label{tab:novel_view_sythesis}
\end{table}

In dietary assessment applications, participants are expected to take minimal actions when capturing food-related media, such as recording a short video with limited food pose coverage. These applications serve as ideal test grounds for Novel View Synthesis and 3D Mesh Reconstruction algorithms. In this section, we present preliminary results for these two tasks using both video captures and Blender-rendered images. For novel view synthesis, we select one object per category and apply recent algorithms, Nerfacto \cite{nerfstudio} and Gaussian Splatting (GS) \cite{gaussiansplatting}, using their official code under default settings. The models are trained on 90\% of the data and tested on the remaining 10\%. We follow \cite{mildenhall2021nerf} and report PSNR, SSIM, and LPIPS scores. 
The results are summarized in Table \ref{tab:novel_view_sythesis}.
Upon inspecting the visual results, we observe that Nerfacto struggles with our dataset. In some video-captured scenes, Nerfacto fails to learn the foreground object, resulting in only a pure background color, whereas GS successfully synthesizes all objects.
We further tested the Nerfacto method by providing it with foreground masks. Visually, we observed that the foreground was correctly learned, but this approach created artifacts in the background, leading to poor quantitative results as shown in Table \ref{tab:novel_view_sythesis}. Therefore, masking plays a crucial role for the Nerf-based method, Nerfacto, on video data but not on rendered data. This discrepancy highlights the challenging non-uniform sparse views and object scale variations in our video data.
For 3D mesh reconstruction, we apply Nerfacto with surface normal prediction settings. Poisson surface reconstruction is then applied to the trained Nerfacto model to obtain the reconstructed mesh. The predicted object meshes from rendered images are compared to the original meshes using Chamfer distance (CD). However, 5 out of 131 objects fail to reconstruct, while the remaining meshes have an average CD of 848.54. For video data, we only provide one of the qualitative results in Figure \ref{fig:nerfacto-nacho} due to the labor-intensive process of pose alignment with the scanned ground truth object. These results underscore the challenging nature of our dataset.




\begin{figure}[t] 
  \centering
    \includegraphics[width=0.9\columnwidth]{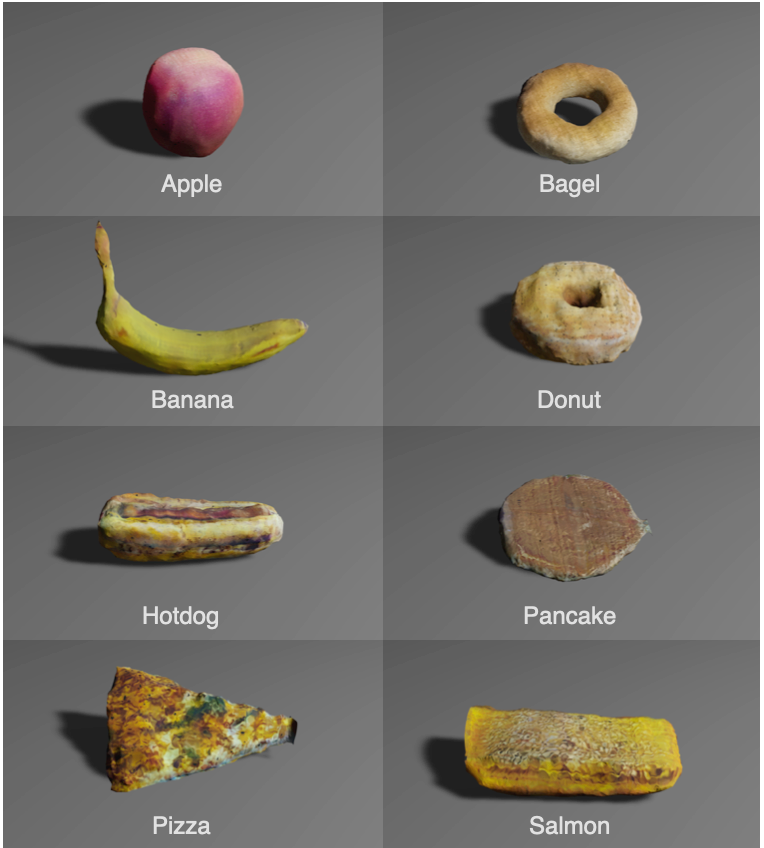}
    \caption{\textbf{MetaFood3D utilizes GET3D~\cite{gao2022get3d} to generate a diverse array of food objects.}}
    \label{fig:3d_gen}
    \vspace{-0.4cm}
\end{figure}

\begin{table}[t!]
\centering
\resizebox{1.0\columnwidth}{!}{
\begin{tabular}{l|c|c|c|c}
\toprule
\multirow{2}{*}{Food object} & Volume & Energy Estimate & \multirow{2}{*}{FID ($\downarrow$)} & \multirow{2}{*}{CD x $10^{3}$ ($\downarrow$)} \\ 
                             & (cm$^{3}$) & (kCal) & & \\ \midrule
Apple                 & 278.88    &217.36 & 105.55  & 5.45  \\
Bagel                 & 326.04   &308.58  & 129.01  & 58.48  \\ 
Banana                & 274.69   &260.01  & 94.26    & 10.75\\
Donuts                & 315.03    &578.60 & 93.15 &   4.44 \\
Hotdog                & 898.01   &501.44 & 99.81  & 4.69  \\
Pancake               &358.24   & 1205.26 & 106.11 &   42.60 \\
Pizza                 &129.83 & 186.37  & 76.09  &  7.10 \\
Salmon        & 202.20   & 573.98 & 108.19     & 18.56 \\
\bottomrule

\end{tabular}
}
\caption{\textbf{Qualitative results for different generated food objects} with volume and energy estimates}
\label{tab:3d_gen_qualitative}

\end{table}

\begin{figure*}[t]
  \centering
    \includegraphics[width=2\columnwidth, bb=0 0 1087 356]{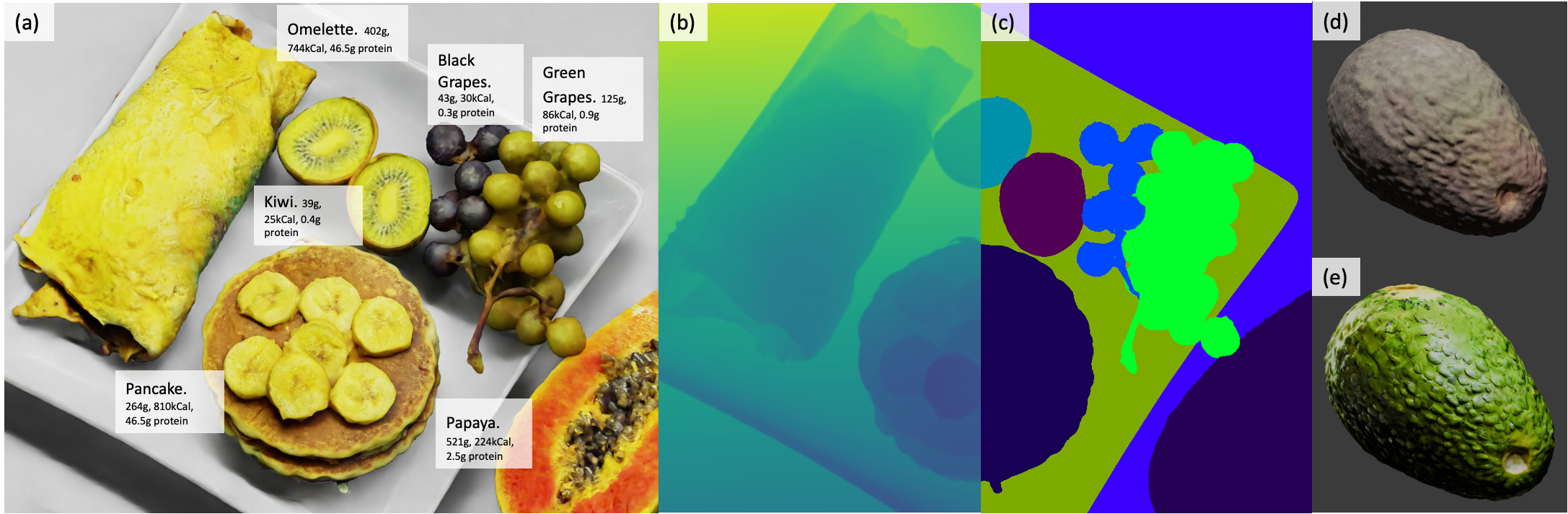}
    \caption{\textbf{(a)} \textbf{Synthetic scene generation} in NVIDIA Omniverse, composed using individual food objects from MetaFood3D. This scene displays a breakfast plate with associated nutrition values for each item including a total weight of 1,433g, 1,944kCal energy, 70g protein, 103g fat, and 191g carbs. \textbf{(b)} Depth map.   \textbf{(c)} Instance segmentation mask. \textbf{(d)} 3D model of an avocado from MetaFood3D, characterized by a brown and dull skin texture. \textbf{(e)} The same avocado mesh as in (d), enhanced with a new texture file generated using Text2Tex \cite{text2tex} with the prompt: \textit{avocado}.}
    \label{fig:synthetic_scene}
    \vspace{-0.6cm}
\end{figure*}

\subsection{Food Scene Synthesis and 3D Food Generation}
\label{subsec: 3d food generation and rendering}
One of the major challenges in food computing, particularly in food portion estimation and nutritional value assessment, is the lack of ground-truth data with precise volume and nutritional measurements for most food datasets \cite{food101, food_rec_2023}. Datasets \cite{nutrition5k, ecustfd_food_dataset} that do include nutritional information often lack diversity in camera perspectives and food combinations, limiting their effectiveness for training robust models. Collecting datasets with diverse view settings and food combinations is costly due to the expense of purchasing food, time-consuming because of the need for precise weighing, and complex because of capturing multiple camera angles, making it difficult to scale. 
Inspired by the highly successful sim-to-real approaches in robotics \cite{metagraspnetv2} and autonomous driving \cite{pun2023neural}, MetaFood3D was developed to address these challenges by providing 3D food objects for diverse eating occasion simulations. These simulations render diverse eating occasion images along with corresponding ground-truth data, including precise nutritional values and portion sizes, which facilitate the development of large-scale, diverse, and realistic datasets for training food computing models. Additionally, this approach can be enhanced with advanced texture generation and 3D food object generation, further increasing the diversity of eating occasion simulations. The following paragraphs present our results in food scene synthesis and 3D food object generation.

\textbf{Food Scene Synthesis.} MetaFood3D supports the creation of synthetic eating scenes with adjustable parameters such as food item placement, portion sizes, and nutrition composition. As shown in Figure \ref{fig:synthetic_scene} (a)(b)(c), we create a breakfast scene in NVIDIA Omniverse simulation engine~\cite{issac-sim}, complete with ground truth labels such as nutrition values, segmentation masks, and depth map. Additionally, the ground truth of bounding boxes and object 6D poses can also be extracted. These scenes can be automatically generated with realistic physics-accurate object interactions in the simulation. Furthermore, texture generation techniques \cite{text2tex} can be leveraged to augment food appearances as shown in Figure \ref{fig:synthetic_scene} (d)(e).

\textbf{3D food object generation.} We use GET3D~\cite{gao2022get3d} to generate textured 3D meshes for various food categories in our dataset. We train the GET3D model from scratch for each selected food type separately, using 3,500 epochs and an average of 750 rendered images per object at a resolution of 512. To compensate for the smaller initial object count compared to the dataset used in GET3D, we set the gamma value to 3,000, penalizing the discriminator and encouraging the generation of more realistic meshes.
We demonstrate the quality of the generated objects through FID~\cite{GAN-FID} and Chamfer Distance (CD)\cite{chamfer_distance_barrow} as shown in Table \ref{tab:3d_gen_qualitative}. A unique aspect of our 3D generation is the inclusion of volume and energy estimates for each generated food object. The energy estimates are calculated based on the generated object's volume, determined using Blender, and the corresponding FNDDS food codes provided by our dataset's nutrition values. 
This enhances the realism of the generated objects, enables accurate energy calculations, and improves dietary assessment functionalities. 
Figure~\ref{fig:3d_gen} visualizes our 3D generation that feature natural textures and coherent shapes enriched by geometric details.

\subsection{Food Portion Estimation}
\label{subsec: portion estimation}
The food portion estimation is a challenging yet important task for food image analysis. Leveraging the rich nutrition value annotations and 3D information in the MetaFood3D dataset, 
we compare the performance of different portion estimation methods covering the four major approaches (stereo-based, depth-based, model-based, and neural network-based) as discussed in Section~\ref{sec:realted_work}. Specifically, we sample 2 frames from the captured video for each food item in the dataset. The food items are divided into training and testing sets, with one food item per category in the testing set and the remaining items in the training set. Overall, the training set contains 1,036 images, while the testing set consists of 216 images. All methods are evaluated on the same testing set for a fair comparison. We compare the methods using Mean Absolute Error (MAE) and Mean Absolute Percentage Error (MAPE). 
We use V-MAE and V-MAPE for volume estimation (cm$^{3}$), and E-MAE and E-MAPE for energy estimation (kCal). Neural network-based methods directly regress energy values, so V-MAE and V-MAPE are not available for them.
\newcolumntype{Y}{>{\centering\arraybackslash}X}

\begin{table}[h]
    \resizebox{\columnwidth}{!}{
    \begin{tabular}{l|cccc}
        \toprule
        \textbf{Method} & \textbf{V-MAE} & \textbf{V-MAPE}  &\textbf{E-MAE} & \textbf{E-MAPE} \\
        \midrule
        Baseline & 165.75 & 836.50 & 214.55 & 1135.93 \\
        Stereo Reconstruction \cite{Dehais2017TwoView} & 153.58 & 214.95 & 262.07 & 244.80  \\
        Voxel Reconstruction \cite{Fang2016Depth} & 120.16 & 96.31  & 174.45 & 130.16 \\
        RGB Only \cite{Shao2021EnergyDensity} & - & - & 1500.23 & 370.9  \\
        Density Map Only \cite{Vinod2022EnergyDensityDepth} & -&- & 1098.87 & 654.33  \\
        Density Map Summing \cite{Ma2023DensityMapSumming} & -&- & 426.68 & 146.18 \\
        \midrule
        3D Assisted Portion \cite{vinod2024Model3D} & 186.45 & 83.26 & 287.11 &  132.42 \\
        MPF3D~\cite{jinge2024MPF3D} & \textbf{62.60} & \textbf{41.43}  & \textbf{77.98} &  \textbf{68.05} \\
        \bottomrule
    \end{tabular}
    }
    \caption{\textbf{Comparison of image-based dietary assessment methods on the MetaFood3D dataset.} The last couple of rows are methods that utilize the 3D models in the MetaFood3D dataset for portion estimation }
    \label{tab:portion_estimation}
\end{table}

The results presented in Table~\ref{tab:portion_estimation} highlight the performance of different classes of existing methods on our MetaFood3D dataset. 
The MPF3D~\cite{jinge2024MPF3D} demonstrates the importance of 3D information for portion estimation outperforming stereo-based, depth-based, and network-based methods on all metrics. The 3D Assisted Portion Estimation method \cite{vinod2024Model3D} achieves the second lowest V-MAPE and E-MAPE. The performance improvement offered by the 2 methods of portion estimation that utilize 3D information from our dataset underscores the important role that 3D food models play in the field of food portion estimation. Thus, the MetaFood3D dataset provides a valuable resource for developing and evaluating various dietary assessment techniques.

\section{Conclusion}
\label{sec:conclusion}
In this paper, we present MetaFood3D, a food-specific 3D object dataset to advance food computing and 3D computer vision. This new dataset provides a robust benchmark for developing and evaluating 3D vision algorithms for real-world scenarios. The dataset features diverse intra-class variations, detailed nutrition annotations and rich multimodal data.
Experimental results demonstrate the strong capabilities of our dataset in food portion estimation, synthetic eating occasion simulation, and 3D food object generation.

{
    \small
    \bibliographystyle{ieeenat_fullname}
    \bibliography{main}
}


\end{document}